\documentclass{article}

\PassOptionsToPackage{numbers, compress}{natbib}
\usepackage[final]{mlsafety_neurips_2022}

\usepackage[utf8]{inputenc} % allow utf-8 input
\usepackage[T1]{fontenc}    % use 8-bit T1 fonts
\usepackage{hyperref}       % hyperlinks
\usepackage{url}            % simple URL typesetting
\usepackage{booktabs}       % professional-quality tables
\usepackage{amsfonts}       % blackboard math symbols
\usepackage{nicefrac}       % compact symbols for 1/2, etc.
\usepackage{microtype}      % microtypography
\usepackage{xcolor}         % colors
\usepackage{amsmath}
\usepackage{subfigure}
\usepackage{graphicx}

\DeclareMathOperator*{\argmax}{arg\,max}
\DeclareMathOperator*{\argmin}{arg\,min}

\title{An Efficient Framework for Monitoring Subgroup Performance of Machine Learning Systems}

\author{%
  Huong Ha \\
  School of Computing Technologies \\
  RMIT University\\
  Melbourne, Australia \\
  \texttt{huong.ha@rmit.edu.au} \\
}

\begin{document}

\maketitle

\global\long\def\se{\hat{\text{se}}}%

\global\long\def\interior{\text{int}}%

\global\long\def\boundary{\text{bd}}%

\global\long\def\new{\text{*}}%

\global\long\def\stir{\text{Stirl}}%

\global\long\def\dist{d}%

\global\long\def\HX{\entro\left(X\right)}%
 
\global\long\def\entropyX{\HX}%

\global\long\def\HY{\entro\left(Y\right)}%
 
\global\long\def\entropyY{\HY}%

\global\long\def\HXY{\entro\left(X,Y\right)}%
 
\global\long\def\entropyXY{\HXY}%

\global\long\def\mutualXY{\mutual\left(X;Y\right)}%
 
\global\long\def\mutinfoXY{\mutualXY}%

\global\long\def\xnew{y}%

\global\long\def\bm{\mathbf{m}}%

\global\long\def\bx{\mathbf{x}}%

\global\long\def\bw{\mathbf{w}}%

\global\long\def\bz{\mathbf{z}}%

\global\long\def\bu{\mathbf{u}}%

\global\long\def\bs{\boldsymbol{s}}%

\global\long\def\bk{\mathbf{k}}%

\global\long\def\bX{\mathbf{X}}%

\global\long\def\tbx{\tilde{\bx}}%

\global\long\def\by{\mathbf{y}}%

\global\long\def\bY{\mathbf{Y}}%

\global\long\def\bZ{\boldsymbol{Z}}%

\global\long\def\bU{\boldsymbol{U}}%

\global\long\def\bv{\boldsymbol{v}}%

\global\long\def\bn{\boldsymbol{n}}%

\global\long\def\bh{\boldsymbol{h}}%

\global\long\def\bV{\boldsymbol{V}}%

\global\long\def\bK{\mathbf{K}}%

\global\long\def\bbeta{\gvt{\beta}}%

\global\long\def\bmu{\gvt{\mu}}%

\global\long\def\btheta{\boldsymbol{\theta}}%

\global\long\def\blambda{\boldsymbol{\lambda}}%

\global\long\def\bgamma{\boldsymbol{\gamma}}%

\global\long\def\bpsi{\boldsymbol{\psi}}%

\global\long\def\bphi{\boldsymbol{\phi}}%

\global\long\def\bpi{\boldsymbol{\pi}}%

\global\long\def\eeta{\boldsymbol{\eta}}%

\global\long\def\bomega{\boldsymbol{\omega}}%

\global\long\def\bepsilon{\boldsymbol{\epsilon}}%

\global\long\def\btau{\boldsymbol{\tau}}%

\global\long\def\bSigma{\gvt{\Sigma}}%

\global\long\def\realset{\mathbb{R}}%

\global\long\def\realn{\realset^{n}}%

\global\long\def\integerset{\mathbb{Z}}%

\global\long\def\natset{\integerset}%

\global\long\def\integer{\integerset}%

\global\long\def\natn{\natset^{n}}%

\global\long\def\rational{\mathbb{Q}}%

\global\long\def\rationaln{\rational^{n}}%

\global\long\def\complexset{\mathbb{C}}%

\global\long\def\comp{\complexset}%

\global\long\def\compl#1{#1^{\text{c}}}%

\global\long\def\and{\cap}%

\global\long\def\compn{\comp^{n}}%

\global\long\def\comb#1#2{\left({#1\atop #2}\right) }%

\global\long\def\nchoosek#1#2{\left({#1\atop #2}\right)}%

\global\long\def\param{\vt w}%

\global\long\def\Param{\Theta}%

\global\long\def\meanparam{\gvt{\mu}}%

\global\long\def\Meanparam{\mathcal{M}}%

\global\long\def\meanmap{\mathbf{m}}%

\global\long\def\logpart{A}%

\global\long\def\simplex{\Delta}%

\global\long\def\simplexn{\simplex^{n}}%

\global\long\def\dirproc{\text{DP}}%

\global\long\def\ggproc{\text{GG}}%

\global\long\def\DP{\text{DP}}%

\global\long\def\ndp{\text{nDP}}%

\global\long\def\hdp{\text{HDP}}%

\global\long\def\gempdf{\text{GEM}}%

\global\long\def\ei{\text{EI}}%

\global\long\def\rfs{\text{RFS}}%

\global\long\def\bernrfs{\text{BernoulliRFS}}%

\global\long\def\poissrfs{\text{PoissonRFS}}%

\global\long\def\grad{\gradient}%
 
\global\long\def\gradient{\nabla}%

\global\long\def\cpr#1#2{\Pr\left(#1\ |\ #2\right)}%

\global\long\def\var{\text{Var}}%

\global\long\def\Var#1{\text{Var}\left[#1\right]}%

\global\long\def\cov{\text{Cov}}%

\global\long\def\Cov#1{\cov\left[ #1 \right]}%

\global\long\def\COV#1#2{\underset{#2}{\cov}\left[ #1 \right]}%

\global\long\def\corr{\text{Corr}}%

\global\long\def\sst{\text{T}}%

\global\long\def\SST{\sst}%

\global\long\def\ess{\mathbb{E}}%

\global\long\def\Ess#1{\ess\left[#1\right]}%

\global\long\def\fisher{\mathcal{F}}%

\global\long\def\bfield{\mathcal{B}}%
 
\global\long\def\borel{\mathcal{B}}%

\global\long\def\bernpdf{\text{Bernoulli}}%

\global\long\def\betapdf{\text{Beta}}%

\global\long\def\dirpdf{\text{Dir}}%

\global\long\def\gammapdf{\text{Gamma}}%

\global\long\def\gaussden#1#2{\text{Normal}\left(#1, #2 \right) }%

\global\long\def\gauss{\mathbf{N}}%

\global\long\def\gausspdf#1#2#3{\text{Normal}\left( #1 \lcabra{#2, #3}\right) }%

\global\long\def\multpdf{\text{Mult}}%

\global\long\def\poiss{\text{Pois}}%

\global\long\def\poissonpdf{\text{Poisson}}%

\global\long\def\pgpdf{\text{PG}}%

\global\long\def\wshpdf{\text{Wish}}%

\global\long\def\iwshpdf{\text{InvWish}}%

\global\long\def\nwpdf{\text{NW}}%

\global\long\def\niwpdf{\text{NIW}}%

\global\long\def\studentpdf{\text{Student}}%

\global\long\def\unipdf{\text{Uni}}%

\global\long\def\transp#1{\transpose{#1}}%
 
\global\long\def\transpose#1{#1^{\mathsf{T}}}%

\global\long\def\mgt{\succ}%

\global\long\def\mge{\succeq}%

\global\long\def\idenmat{\mathbf{I}}%

\global\long\def\trace{\mathrm{tr}}%

\global\long\def\argmax#1{\underset{_{#1}}{\text{argmax}} }%

%\global\long\def\argmin#1{\underset{_{#1}}{\text{argmin}\ } }%

\global\long\def\diag{\text{diag}}%

\global\long\def\norm{}%

\global\long\def\spn{\text{span}}%

\global\long\def\vtspace{\mathcal{V}}%

\global\long\def\field{\mathcal{F}}%
 
\global\long\def\ffield{\mathcal{F}}%

\global\long\def\inner#1#2{\left\langle #1,#2\right\rangle }%
 
\global\long\def\iprod#1#2{\inner{#1}{#2}}%

\global\long\def\dprod#1#2{#1 \cdot#2}%

\global\long\def\norm#1{\left\Vert #1\right\Vert }%

\global\long\def\entro{\mathbb{H}}%

\global\long\def\entropy{\mathbb{H}}%

\global\long\def\Entro#1{\entro\left[#1\right]}%

\global\long\def\Entropy#1{\Entro{#1}}%

\global\long\def\mutinfo{\mathbb{I}}%

\global\long\def\relH{\mathit{D}}%

\global\long\def\reldiv#1#2{\relH\left(#1||#2\right)}%

\global\long\def\KL{KL}%

\global\long\def\KLdiv#1#2{\KL\left(#1\parallel#2\right)}%
 
\global\long\def\KLdivergence#1#2{\KL\left(#1\ \parallel\ #2\right)}%

\global\long\def\crossH{\mathcal{C}}%
 
\global\long\def\crossentropy{\mathcal{C}}%

\global\long\def\crossHxy#1#2{\crossentropy\left(#1\parallel#2\right)}%

\global\long\def\breg{\text{BD}}%

\global\long\def\lcabra#1{\left|#1\right.}%

\global\long\def\lbra#1{\lcabra{#1}}%

\global\long\def\rcabra#1{\left.#1\right|}%

\global\long\def\rbra#1{\rcabra{#1}}%

\begin{abstract}
Monitoring machine learning systems post deployment is critical to ensure the reliability of the systems. Particularly importance is the problem of monitoring the performance of machine learning systems across all the data subgroups (subpopulations). In practice, this process could be prohibitively expensive as the number of data subgroups grows exponentially with the number of input features, and the process of labelling data to evaluate each subgroup's performance is costly. In this paper, we propose an efficient framework for monitoring subgroup performance of machine learning systems. Specifically, we aim to find the data subgroup with the worst performance using a limited number of labeled data. We mathematically formulate this problem as an optimization problem with an expensive black-box objective function, and then suggest to use Bayesian optimization to solve this problem. Our experimental results on various real-world datasets and machine learning systems show that our proposed framework can retrieve the worst-performing data subgroup effectively and efficiently.
\end{abstract}

\section{Introduction} \label{sec:intro}

Supervised machine learning (ML) systems are increasingly deployed in critical application domains but there is no guarantee that these systems will continue to perform well after deployment ~\cite{Sawade2010ActiveTest}. Monitoring the performance of ML systems is thus crucial to prevent potential failures that may cause severe unintended consequences \cite{Ginart2022mlmonitor, Shankar2022OperationML}. Particularly importance is to monitor the performance of ML systems across all the data subgroups so as to discover the scenarios when the systems may show anomalous or faulty behaviours \cite{Shankar2022OperationML}. In practice, there have been various issues reported regarding the malicious performance of well-known ML systems on some particular data subgroups. For example, researchers have noted that in COMPAS \cite{Angwin2016Compas}, an ML system used across US courtrooms to predict future crimes, females who initially committed misdemeanors have their recidivism risk significantly over-estimated for half of the COMPAS risk groups \cite{Zhang2016bias}. Another example is that various commercial facial detection algorithms have been shown to perform significantly different across different data subgroups of races and genders with a worse performance on females with darker skins \cite{Buolamwini2018GenderBias}. In practice, it is recognized that organizations often want to understand the performance of the ML systems on different demographic data subgroups, customer segments, etc. \cite{Shankar2022OperationML}. \textit{Therefore, it is important to monitor the performance of an ML system across the data subgroups (subpopulations)}.

Existing literature tackles this problem by focusing on some limited pre-defined data subgroups (e.g., gender and race). However, this does not include all the possible subgroups, and therefore, important defective subgroups might not be discovered. In this work, we aim to monitor the ML systems' performance across all the possible data subgroups. In particular, we aim to find the data subgroup where an ML system performs the worst. An exhaustive search across all the subgroups could be prohibitively expensive, and in many cases, is impossible. The reason is that evaluating the performance of an ML system on a data subgroup requires the collection of an amount of labelled data corresponding to that subgroup. For example, to evaluate the performance of an ML system for the data subgroup female + black + married, a set of labelled data with input attributes female, black, and married need to be collected. However, the number of possible data subgroups scales exponentially with the number of input attributes (to be used to create subgroups), and thus the amount of labelled data to be collected could be exploded if an exhaustive search is employed. For a dataset with $n$ input attributes (to be used to create subgroups) and each attribute has $m_i$ distinct values, the number of possible data subgroups is $\prod_{i=1}^n m_i$. If it is required to collect $K$ labelled data points to estimate the performance of a data subgroup, then in total, $K\prod_{i=1}^n m_i$ data points need to be labelled, which is expensive and could be impossible in many situations. 

In this paper, we propose an efficient framework that can find the worst-performing data subgroup with a minimal number of labelled data. We first formulate the problem as an optimization problem with an expensive black-box objective function. We then suggest to use Bayesian Optimization (BO), a powerful sequential global optimization technique, so as to find the worst-performing data subgroup efficiently. The key idea of BO is to train a surrogate model from an initial labelled dataset, construct an acquisition function from this surrogate model, and use this function to select the most informative data subgroup(s) that can help to find the worst-performing subgroup most efficiently. Our framework can work with various types of supervised ML systems, e.g. classifier and regressor, and can also work with different performance metrics such as accuracy, mean square error, precision, recall.

% Existing literature has targeted this problem \cite{Kearns2018auditml, Zhang2016bias}, but only tackle few small problems, like the subgroups are pre-defined, and the corresponding fairness related metrics are proposed. Most of the works only targeted several types of metrics or a specific type of machine learning model like classifier. 

% Our proposed framework can find the subgroup with the worst performance using a limited number of labeled data. It can also provide a list of subgroups that perform worse than a user-specified threshold. In practice, the problem of monitoring subgroup performance is challenge because of the number of possible subgroups can be exploded. Even for a small number of attributes (n) with each attribute having $n_i$ categories, the total number of subgroups is the product of all these categories, and this value could be prohibitively large. Besides, note that for each particular subgroup, if we want to estimate its corresponding performance, we need to collect an M number of data points. But labelled data is generally expensive. The goal is to find the bad subgroups with only a limited number of data points so as to efficiently monitoring the system.

In summary, our contributions are:
\begin{enumerate}
    \item A generic and efficient framework for monitoring the subgroup performance of ML systems;
    \item An effective technique that can help to find the worst-performing data subgroup using a minimal number of labelled data;
    \item A set of experiments to demonstrate the efficacy of our proposed framework.
\end{enumerate}

% \vspace{-1mm}
\paragraph{Related Work} 
There have been various research works tackling the problem of assessing or monitoring the subgroup performance of an ML system, however, their settings are generally different compared to ours. Please refer to Section \ref{sec:related-work} in the Appendix for more details.

% \vspace{-1mm}
\section{An Efficient Framework for Monitoring Subgroup Performance}

\begin{figure*}
  \begin{center}
  \includegraphics[trim=0cm 0cm 0cm  0cm, clip, width=0.37\linewidth,height=39.2mm]{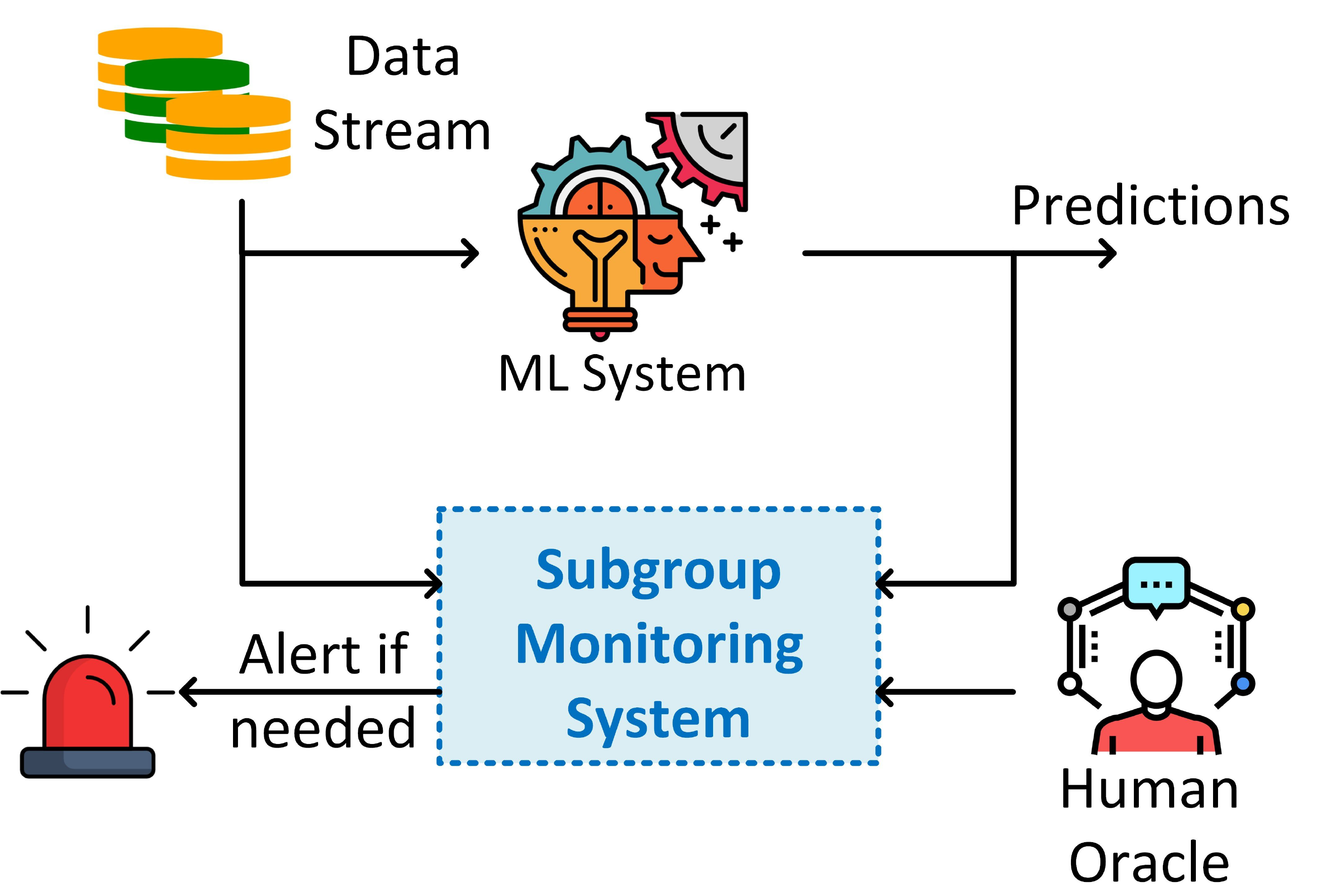}
  \hspace{4mm}
    \includegraphics[trim=0cm 0cm 0cm  0cm, clip, width=0.58\linewidth,height=39.0mm]{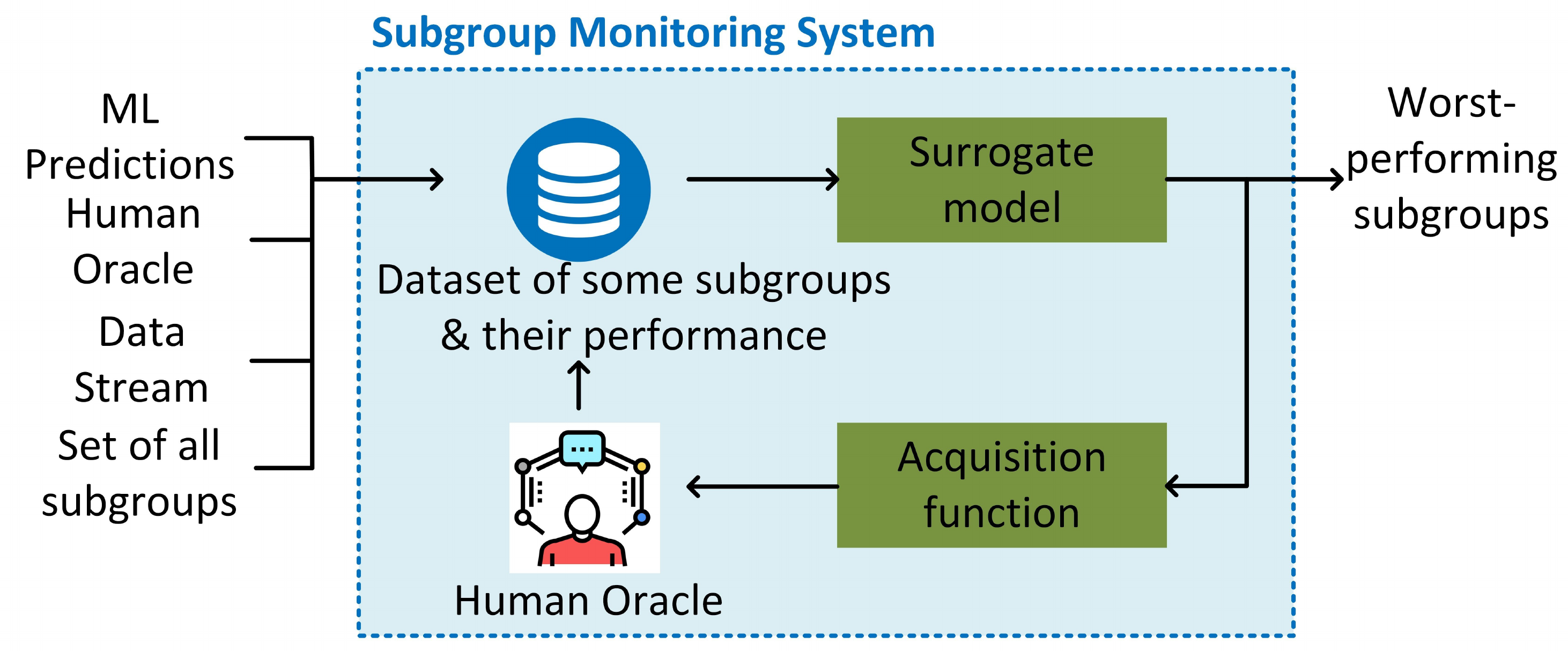}
    \caption{\textit{Left}: The subgroup performance monitoring workflow; the monitoring system takes the data stream, the ML systems' predictions, and a human oracle as inputs, and outputs the worst-performing data subgroup that can alert users regarding the quality of the ML system. \textit{Right}: The detailed implementation of our proposed subgroup performance monitoring framework based on BO method.}
    \label{fig:framework}
    \vspace{-2mm}
 \end{center}
\end{figure*}

In this section, we present our proposed framework for monitoring subgroup performance of an ML system. An overview of our proposed framework is shown in Figure \ref{fig:framework}.

% \vspace{-1mm}
\subsection{Formulating the Subgroup Monitoring Problem}

Let us denote an individual input data and its ground-truth as $((x, x'), y_{x,x'})$ where $x \in \mathcal{X}$ denotes the attributes that are used to construct the data subgroups, $x' \in \mathcal{X}'$ as the attributes of non-interest, and $y_{x,x'}$ as the corresponding ground-truth of $(x,x')$. For example, for a problem of using census data of a person to predict their income, one can set the attributes $x$ to be gender, race, relationship, age whilst other attributes can be set as attributes of non-interest $x'$. The choice of attributes to be used to construct the subgroups depends on the users, in particular, it depends on which data segments the users want to monitor. Given an ML system $C$, let us denote $C(x,x')$ as the prediction of $C$ for an input $(x,x')$. The performance value of the system $C$ with a metric $M$ can then be expressed as,
\begin{equation} \nonumber
    M(C) = M(C_{\mathcal{X}, \mathcal{X}'}, \mathcal{Y}_{\mathcal{X}, \mathcal{X}'}),
\end{equation}
where $C_{\mathcal{X}, \mathcal{X}'} = \{ C(x,x') \}_{x \in \mathcal{X}, x' \in \mathcal{X}'}$, $\mathcal{Y}_{\mathcal{X}, \mathcal{X}'} = \{ y_{x,x'} \}_{x \in \mathcal{X}, x' \in \mathcal{X}'}$. Then the performance value of the ML system $C$ w.r.t. a data subgroup $x$ can be computed as,
\begin{equation} \nonumber
    M_x(C) = M(C_{x, \mathcal{X}'}, \mathcal{Y}_{x, \mathcal{X}'}).
\end{equation}
The problem of finding the worst-performing data subgroup becomes the problem of finding $x^*$:
\begin{equation} \label{eq:problem}
    x^* = \argmin\nolimits_{x \in \mathcal{X}} M_x(C).
\end{equation}

As discussed in Section \ref{sec:intro}, computing $M_x(C)$ for each data subgroup $x$ is expensive, therefore, it is generally impossible to exhaustively search for $x^*$. In this paper, we aim to solve the optimization problem in Eq. (\ref{eq:problem}) in an efficient manner, i.e., using a minimal number of evaluations of $M_x(C)$. 

% Suppose we have the input domain that contains two types of attributes: attributes that can build the subgroups that we care (attributes of interest), and attributes that we do need to specifically understand the performance of the subgroups generated by these attributes. Let us denote an individual input data and its ground-truth as $((x, x'),y)$ where $x \in \mathcal{X}$ denotes the segmentable input features, $x' \in \mathcal{X}'$ as the normal input features, and $y$ as the corresponding ground-truth of $(x,x')$. Given a metric $M$, an ML model $C$, let us denote $M(C)$ is the performance value of the model $C$ with metric $M$. Let us then denote $M_x(C)$ as the performance value of the ML model $C$ corresponding to the subgroup $x$. That is the performance of the ML model when the input features are $x,x'$ and $x'$ is any value in $\mathcal{X}'$.

% In practice, we can estimate the value of $M_x(C)$ for each value of $x$ by collecting a number of labeled data ($N$) corresponding to $x$. However, this process is very expensive. A search only with $10$ tries of $x$ can result in the cost of labelling $10N$ data points. Therefore, it is critical to find the worst subgroup performance within a limited number of trials. Besides, in the deployment setting (online setting), it is also critical to find the worst subgroup performance as quick as possible.

\subsection{An Efficient Subgroup Searching Methodology}

 To solve the optimization problem in Eq. (\ref{eq:problem}) in an efficient manner, we propose to use Bayesian Optimization (BO) technique \cite{Jones1998BO, Jones2001BO, Snoek2012BO, Shahriari2016BO}. BO is a powerful optimization method to find the global optimum of an expensive black-box objective function by sequential queries. Applying BO to solve the problem in Eq. (\ref{eq:problem}) can be done as follows.
 
 Firstly, an initial dataset $D_0 = \{(x_i, M_{x_i}(C))\}_{i=1}^{N_0}$ is constructed from a number of data subgroups $\{ x_i \}_{i=1}^{N_0}$ and their corresponding performance values $\{ M_{x_i}(C) \}_{i=1}^{N_0}$. Secondly, a surrogate model is trained on this initial dataset to approximate the behaviour of the performance function $M_x(C)$. Thirdly, an acquisition function is constructed from this surrogate model to suggest the next most informative data subgroup to be evaluated so as to find the worst-performing subgroup $x^{*}$ fastest. The ML system's performance is then evaluated at this data subgroup and the data pair is then added to $D_0$. The process is conducted repeatedly until the labelling budget is depleted. The worst-performing subgroup $\hat{x}^*$ is then chosen as the worst-performing subgroup from the labelled dataset $D_T$ obtained after the BO process, i.e., $\hat{x}^* = \argmin_{x \in D_T} M_{x}(C)$.

In the BO technique, there are various common choices of surrogate models including the Gaussian Process (GP) \cite{Rasmussen2006GP}, the random forest \cite{Hutter2011SMAC, Breiman2001RF}, the neural network \cite{Snoek2015BONN}. In this work, we use a GP as the surrogate model of the BO process as GP has been shown to be effective in many practical scenarios \cite{Snoek2012BO, Shahriari2016BO}. Note that, in the subgroup performance monitoring problem, the data subgroups $x$ are categorical variables, so we use the corresponding BO technique for categorical variables. In particular, the categorical variables are one-hot encoded, so that they can be treated as continuous variables, and then the standard continuous BO process is performed on the transformed variables. Finally, we use the Expected Improvement (EI) \cite{Mockus1975BOEI} as the acquisition function as it is one of the most effective and well-studied acquisition functions in the BO literature \cite{Shahriari2016BO}. 

\section{Experimental Results}

In this section, we describe in details the experimental evaluation of our proposed framework. Our experiments aim to answer the question of whether our proposed framework can find the worst-performing data subgroup efficiently.

\vspace{-1mm}
\paragraph{Datasets} We evaluate our proposed framework using two real-world datasets from the UC Irvine (UCI) Data Repository\footnote{https://archive.ics.uci.edu}. The first dataset is the \textit{Adult} dataset \cite{Kohavi1996Adult}. Each record in this dataset describes the census data of a person and the goal is to predict whether the income of this person exceeds 50,000 USD per year. We choose the subgroup attributes to be age, race, gender and relationship. The second dataset we use for evaluation is the \textit{Bike Sharing} dataset \cite{Fanaee2013BikeSharing}. This dataset contains the hourly count of rental bikes between years 2011 and 2012 in Capital bikeshare system with the corresponding weather and seasonal information. The goal is to predict the hourly count of rental bikes given a particular weather and seasonal information. For this dataset, we choose the subgroup attributes to be season, weather, hours, and working day.

\vspace{-1mm}
\paragraph{Machine Learning Systems} We split each dataset into two parts: training and testing. We have two choices of training size: 1000 and 2000 data points, the rest of the data is the testing part. We train different ML systems on the training part and then use our proposed framework to find the worst-performing data subgroups of these ML systems on the test part. The \textit{Adult} dataset corresponds to a classification problem, so we use Logistic Regression \cite{Cox1958regression} and Gradient Boosting \cite{Friedman2002GradientBoost} as the ML systems, and the classification accuracy as the performance metric. The \textit{Bike Sharing} dataset corresponds to a regression problem so we use Linear Regression \cite{Hastie2001ElementStatisticLearn} and Gradient Boosting \cite{Friedman2002GradientBoost} as the ML systems, and the mean square error as the performance metric. With two choices of training dataset size, we have in total four ML systems for each dataset. For each combination of dataset and ML system, we repeat the experiments 20 times with different split of training and testing parts, and report the means and standard errors of the worst subgroup performance found at each iteration.

\vspace{-2mm}
\paragraph{Baselines} We compare our proposed framework with two baselines: Random Search (RS) and Exhaustive Search (ES). With RS, we randomly select a number of data subgroups, evaluate the ML system's performance associated with these subgroups, and then retrieve the worst-performing subgroup. With ES, we construct all the possible valid data subgroups, sequentially evaluate the ML system's performance of all the subgroups, and then retrieve the worst-performing subgroup.

\begin{figure*}
  \begin{center}
  \includegraphics[trim=0cm 0cm 0cm  0cm, clip, width=0.24\linewidth]{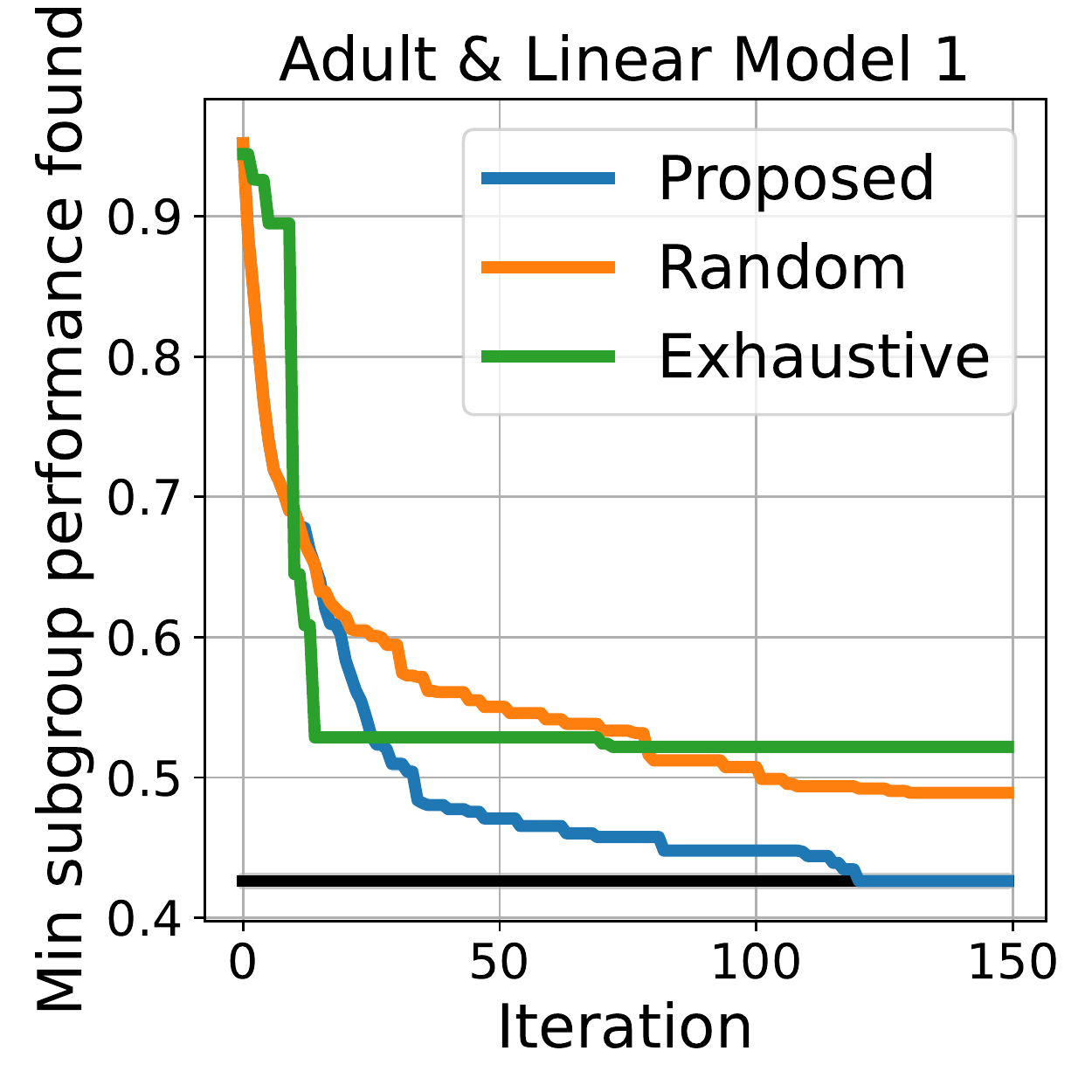}
    \includegraphics[trim=0cm 0cm 0cm  0cm, clip, width=0.24\linewidth]{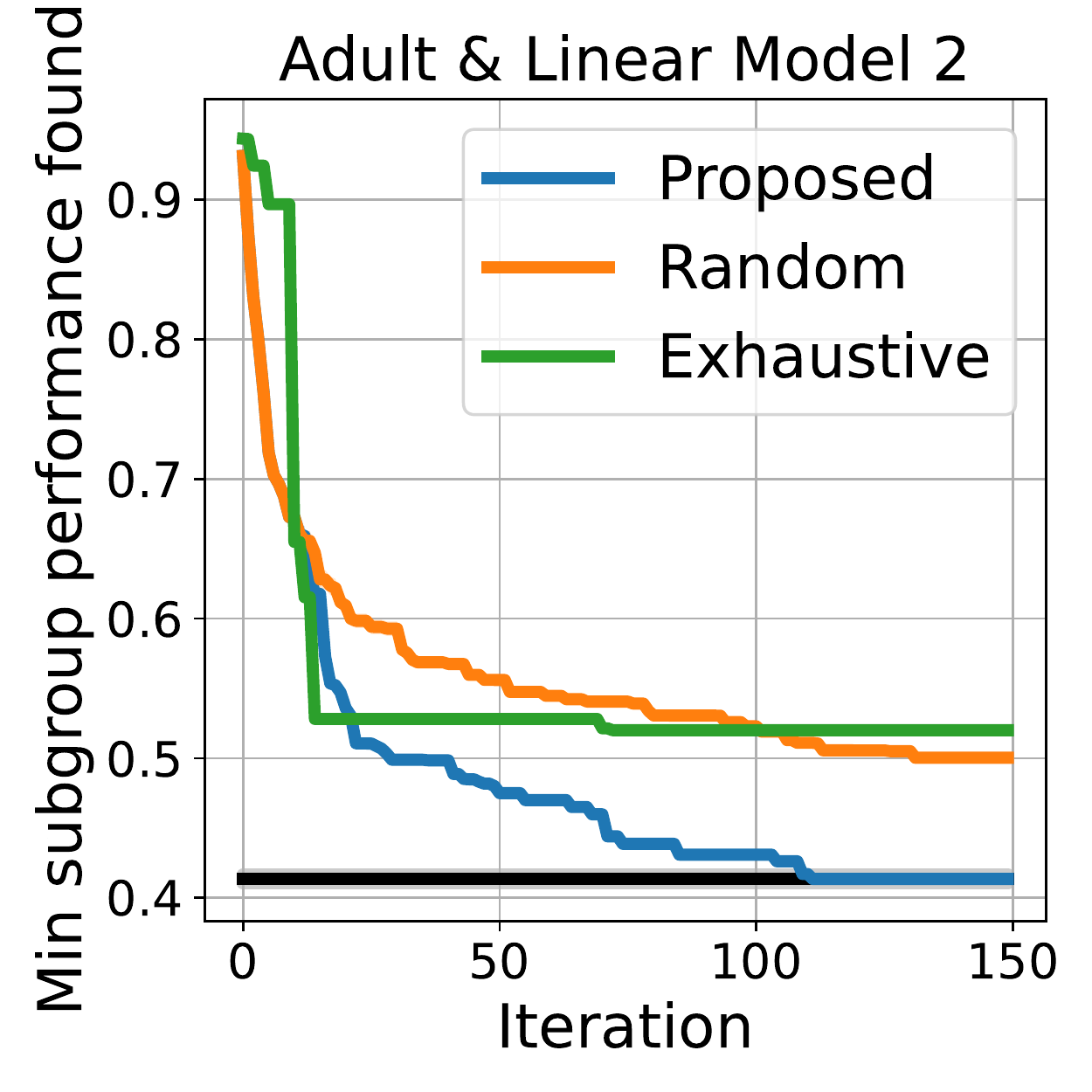}
    \includegraphics[trim=0cm 0cm 0cm  0cm, clip, width=0.24\linewidth]{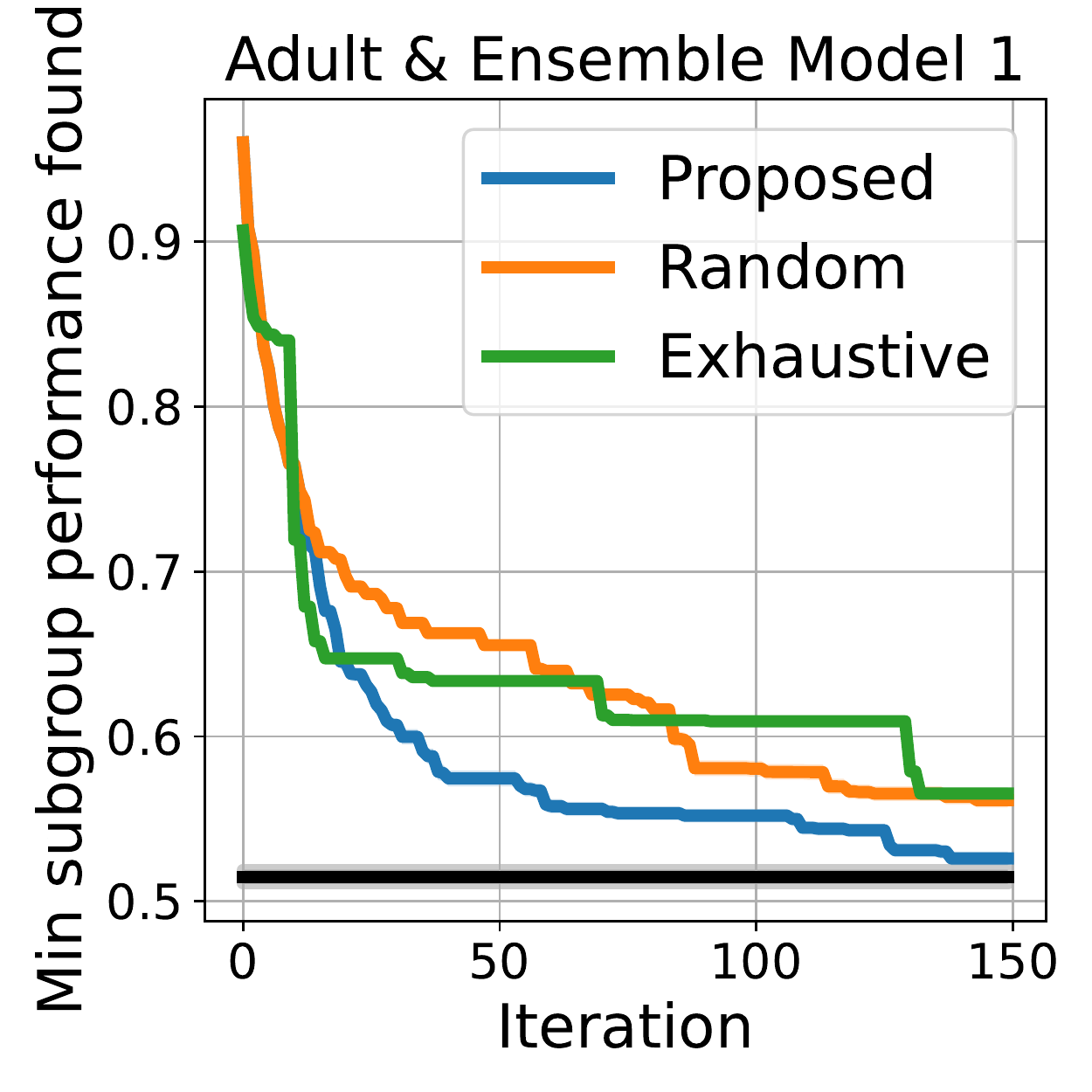}
    \includegraphics[trim=0cm 0cm 0cm  0cm, clip, width=0.24\linewidth]{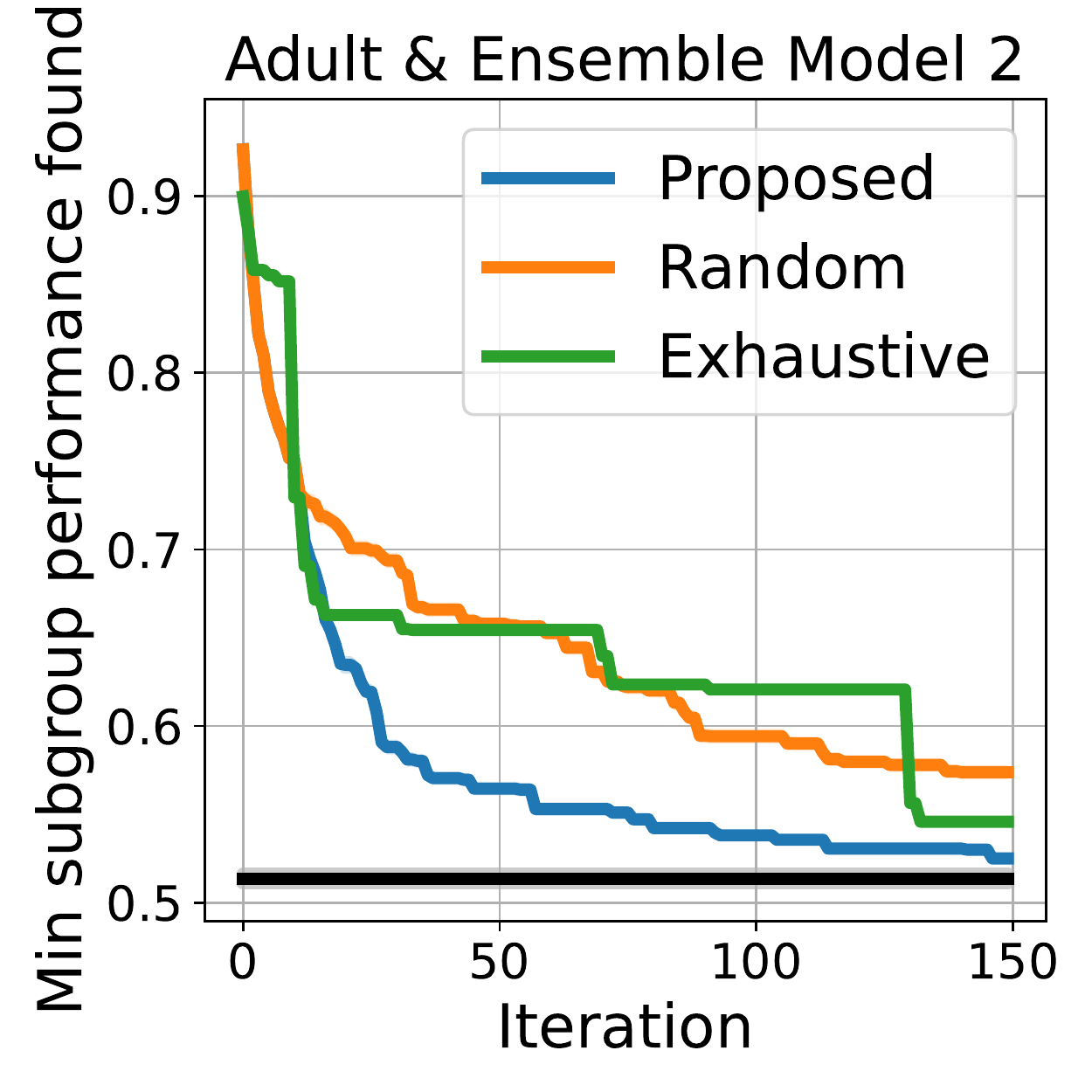}
    \vspace{-2mm}
    \caption{Results of the \textit{Adult} dataset with different ML systems. Black lines are the true worst subgroup performance. Coloured lines are the worst subgroup performance found by the methods.}
    \label{fig:exp-adult}
    \vspace{-1mm}
 \end{center}
\end{figure*}

\begin{figure*}
  \begin{center}
  \vspace{-2mm}
  \includegraphics[trim=0cm 0cm 0cm  0cm, clip, width=0.24\linewidth]{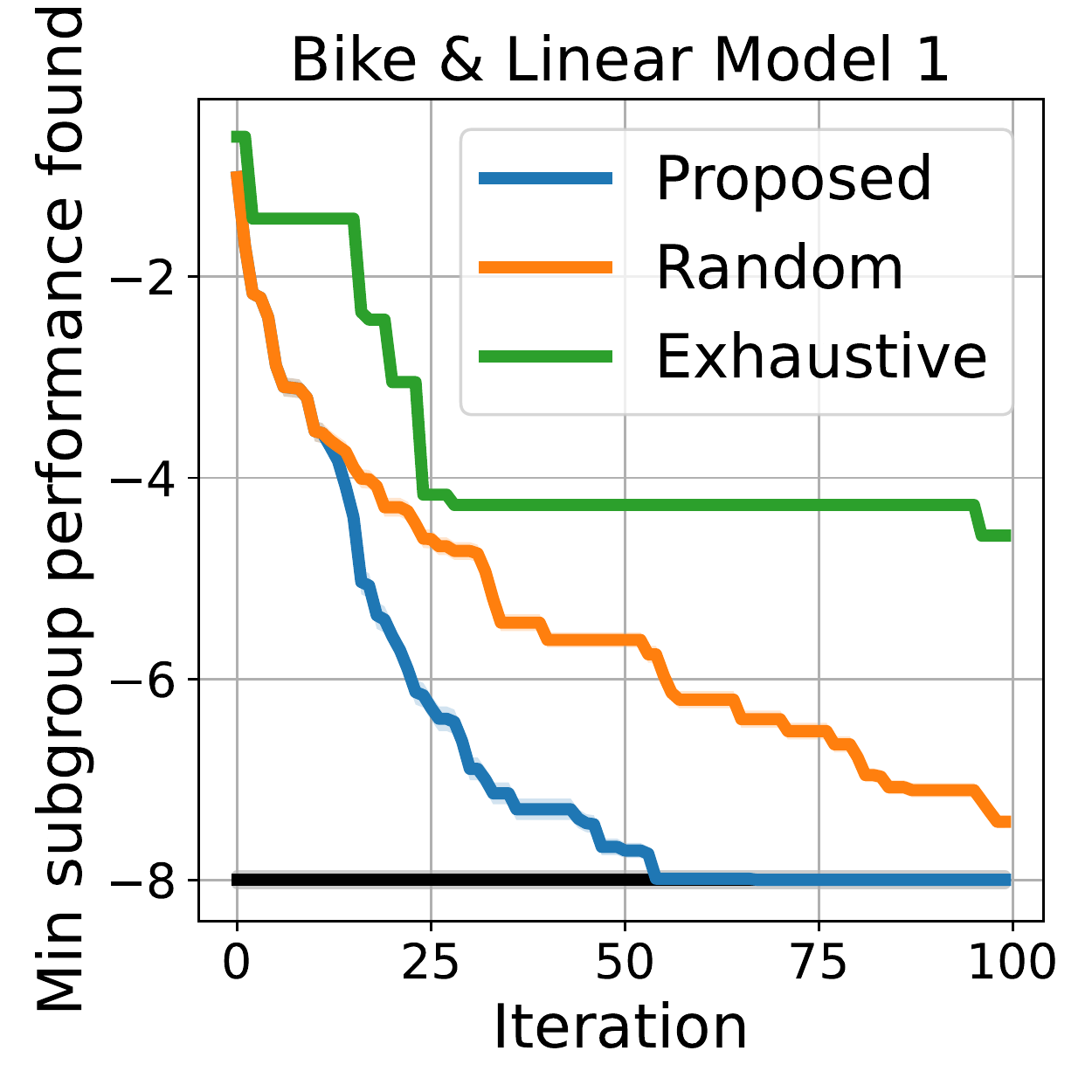}
    \includegraphics[trim=0cm 0cm 0cm  0cm, clip, width=0.24\linewidth]{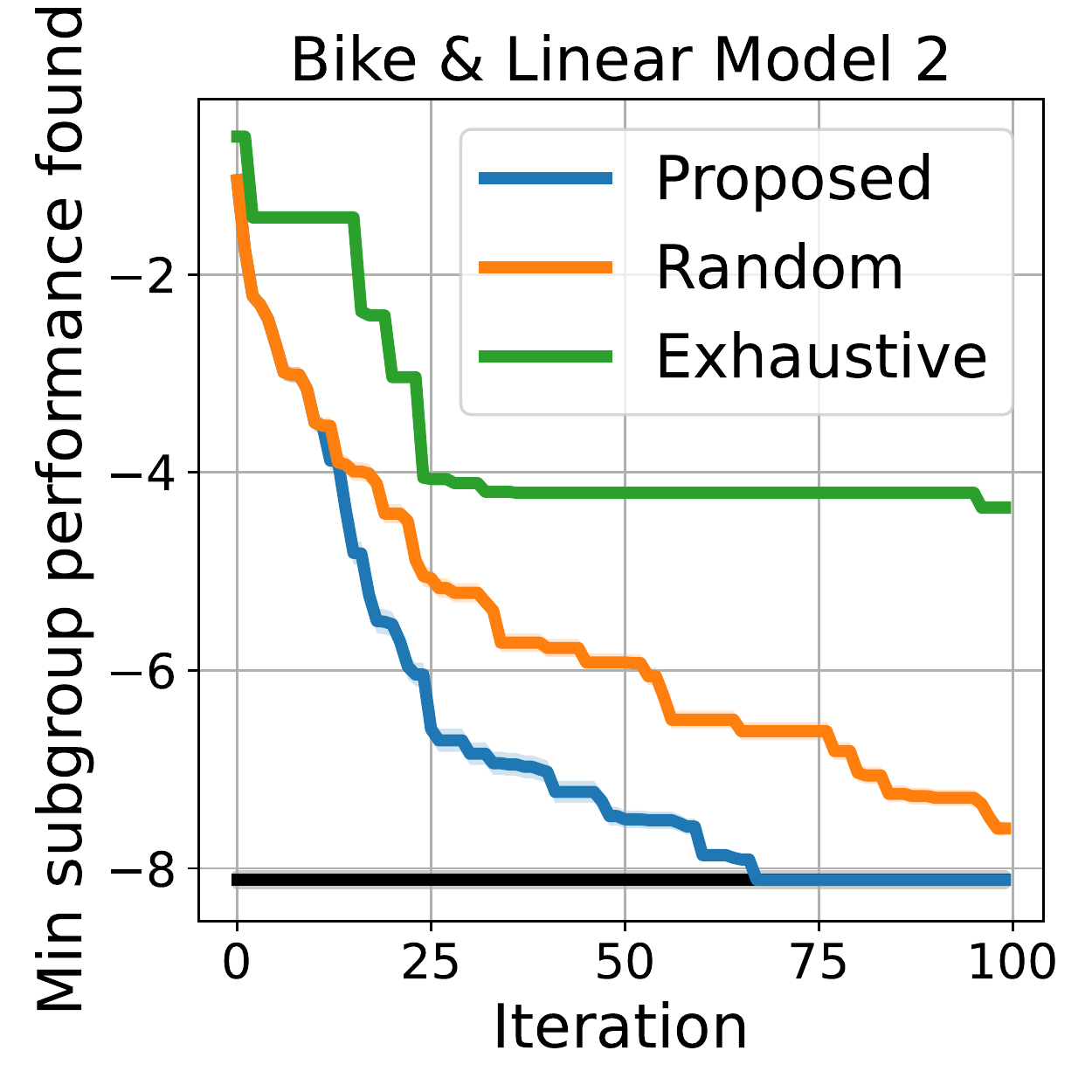}
    \includegraphics[trim=0cm 0cm 0cm  0cm, clip, width=0.24\linewidth]{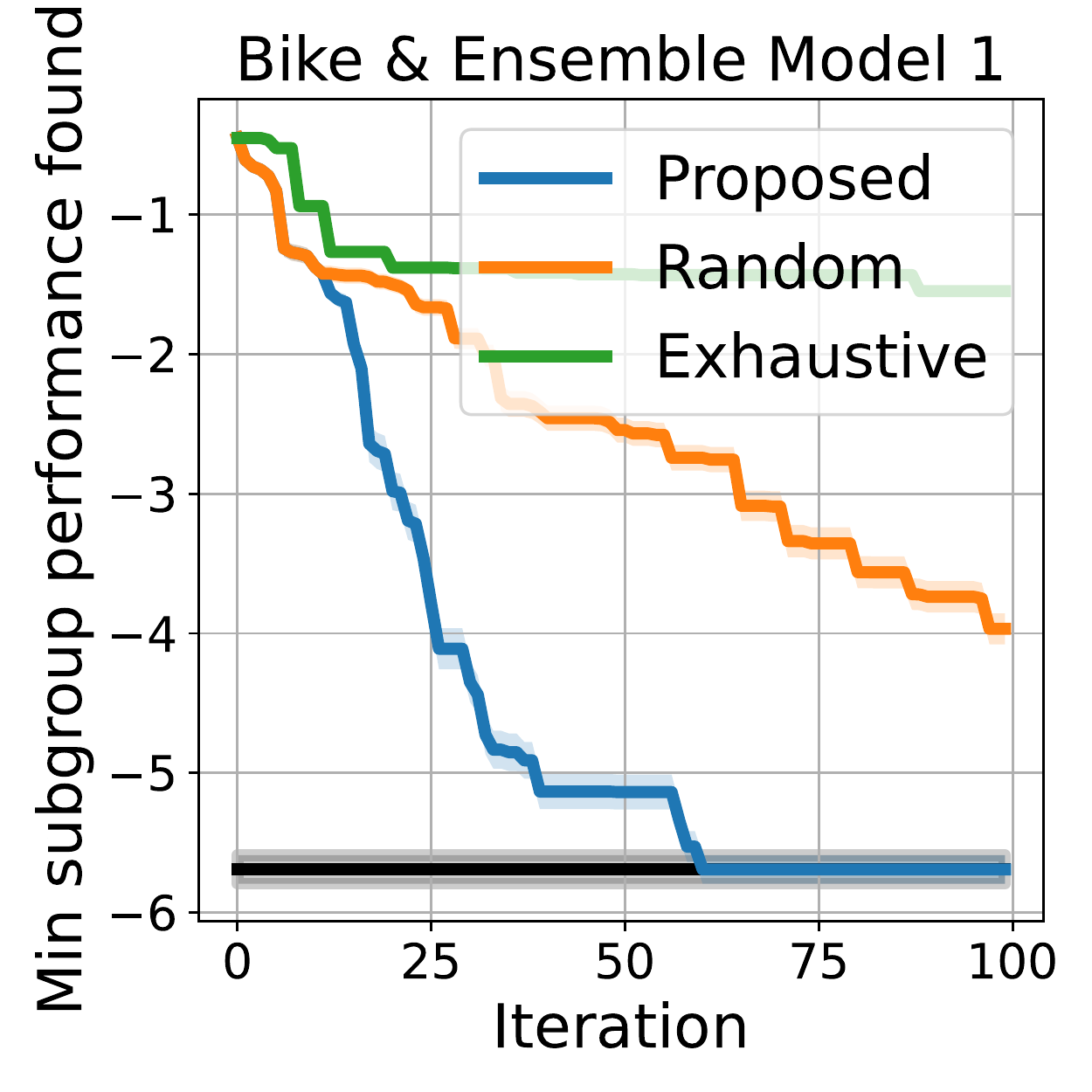}
    \includegraphics[trim=0cm 0cm 0cm  0cm, clip, width=0.24\linewidth]{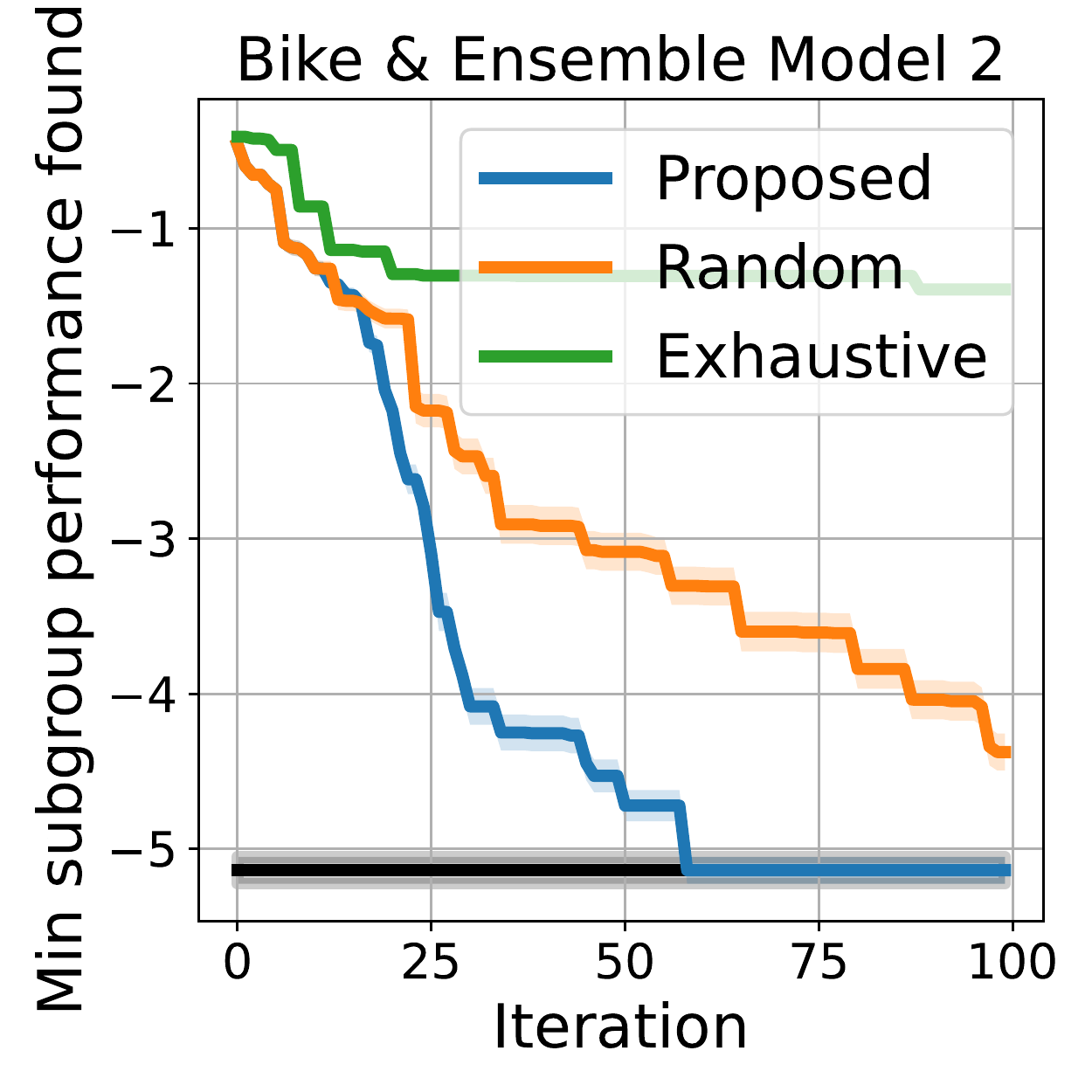}
    \vspace{-2mm}
    \caption{Results of the \textit{Bike Sharing} dataset with different ML systems. Black lines are the true worst subgroup performance. Coloured lines are the worst subgroup performance found by the methods.}
    \label{fig:exp-bike}
    \vspace{-4mm}
 \end{center}
\end{figure*}

In Figures \ref{fig:exp-adult} and \ref{fig:exp-bike}, we show the experimental results of our proposed method and baseline methods for the \textit{Adult} and \textit{Bike Sharing} datasets, respectively. It can be clearly seen that our proposed method can find the worst-performing data subgroup much faster compared to the baseline methods. For the \textit{Adult} dataset, our proposed method can find the worst-performing subgroup after only around 100 iterations (i.e., 100 subgroup performance evaluations) whilst other methods takes much more iterations to find this subgroup. Similarly, for the \textit{Bike Sharing} dataset, our proposed method can find the worst subgroup within 50-70 iterations, which is much faster compared to other baseline methods.

\vspace{-2mm}
\section{Conclusion}
\vspace{-2mm}
In this paper, we have proposed a framework for efficiently monitoring the performance of an ML system across all the data subgroups. In particular, we aim to find the data subgroup with the worst performance. Our proposed framework is data-efficient as it requires a minimal amount of labeled data, and generic as it can be applied to various types of ML systems and performance metrics. Our experimental results on two real-world datasets and various types of ML systems confirm the effectiveness of our proposed framework.

\begin{ack}
The author would like to thank NVIDIA, in particular, the NVIDIA Academic Hardware Grant Program, for the computing support on this project.
\end{ack}

\bibliographystyle{plain}
\bibliography{reference}

\appendix

\section{Appendix}

\subsection{Related Work} \label{sec:related-work}

There are various works targeting the problem of detecting the bad performance of an ML system on some data subgroups, especially in the context of evaluating the biasness of an ML system, for example, \cite{Dwork2018Fairness, Foulds2020Fairness, Morina2019Fairness}. However, most of these work only aim to assess the performance of ML systems on some pre-defined data subgroups such as race, gender. Our work, on the other hand, assesses the performance of an ML system across all the data subgroups.

Directly related to our work, there are some works targeting the problem of assessing the performance of an ML system across all the data subgroups, however, these work can only work on a limited types of ML system (e.g., binary classifiers), and  they do not target the problem of reducing the cost of labelling data. The work in \cite{Kearns2018auditml} suggests considering the fairness statistical notions via a large number of subgroups. However, this work only considers the binary classifiers, on the other hand, our proposed framework can work with different types of ML systems including multi-class classifiers and regressors. The paper by Zhang and Neil \cite{Zhang2016bias} presents a subset scan method to detect if a probabilistic binary classifier has statistically significant bias over or under predicting the risk for some subgroups and identify the characteristics of this subgroup. However, similar to \cite{Kearns2018auditml}, this work also only works for binary classifiers, and it also does not focus on the problem of reducing the cost of labelling data. The work in \cite{Pastor2021SubgroupBiased} presented a method for finding data subgroups that behave differently from the overall dataset. However, the method does not tackle the problem of expensive labelled data as our proposed framework. 

There are works targeting the problem of monitoring ML systems \cite{Ginart2022mlmonitor}, but this work do not care about the performance of the ML systems across all the data subgroups. Finally, there are works that aim to test the performance of an ML system in a data-efficient manner \cite{Sawade2010ActiveTest, Gopakumar2018AlgoAssu, Ha2021ALTMAS, Kossen2021ActiveTesting}, however, these works also do not care about the performance across the data subgroups.

\subsection{Background}

\paragraph{Bayesian Optimization}

Bayesian optimization is a powerful optimization method to find the global optimum of an unknown objective function $f(x)$ by sequential queries \cite{Jones1998BO, Jones2001BO, Snoek2012BO, Shahriari2016BO}. First, at time $t$, a surrogate model is used to approximate the behaviour of $f(x)$ using all the current observed data $D_{t-1} = \{(x_i, y_i)\}^n_{i=1}, y_i = f(x_i) + \zeta_i$, where $\zeta_i \sim  \mathcal{N} (0, \sigma^2)$ is the noise. Second, an acquisition function is constructed from the surrogate model that suggests the next point $x_{itr}$ to be evaluated. The objective function is then evaluated at $x_{itr}$ and the new data point $(x_{itr}, y_{itr})$ is added to $D_{t-1}$.  These steps are conducted in an iterative manner to get the best estimate of the global optimum.

\paragraph{Gaussian Process}

A GP defines a probability distribution over functions $f$ under the assumption that any finite subset $\lbrace (\bx_i, f(\bx_i) \rbrace$ follows a normal distribution \cite{Rasmussen_2006gaussian}. Formally, a GP is defined  as $f(\bx)\sim \text{GP}\left(\mu\left(\bx\right),k\left(\bx,\bx'\right)\right)$, where $\mu(\bx)=\mathbb{E}\left[f\left(\bx\right)\right]$ is the mean function and $k(\bx,\bx')=\mathbb{E}\left[(f\left(\bx\right)-\mu\left(\bx\right))(f\left(\bx'\right)-\mu\left(\bx'\right))^{T}\right]$ is the covariance function \cite{Rasmussen_2006gaussian}.

Assuming a zero mean prior $m(\bx)=0$ for simplicity, we have the joint multivariate Gaussian distribution between the observed point $f$ and a new data point $\bx_{\text{new}},f_{\text{new}}=f\left(\bx_{\text{new}}\right)$ as follows,
\begin{align}
\left[\begin{array}{c}
\boldsymbol{f}\\
f_{\text{new}}
\end{array}\right] & \sim\mathcal{N}\left(0,\left[\begin{array}{cc}
\bK & \bk_{\text{new}}^{T}\\
\bk_{\text{new}} & k_{\text{new}}
\end{array}\right]\right),\label{eq:p(f|f*)}
\end{align}
 where $k_{\text{new}}=k\left(\bx_{\text{new}},\bx_{\text{new}}\right)$, $\bk_{\text{new}}=[k\left(\bx_{\text{new}},\bx_{i}\right)]_{\forall i\le N}$
and $\bK=\left[k\left(\bx_{i},\bx_{j}\right)\right]_{\forall i,j\le N}$. Combining Eq. (\ref{eq:p(f|f*)}) with the fact that $p\left(f_{\text{new}}\mid\boldsymbol{f}\right)$ follows a univariate Gaussian distribution $\mathcal{N}\left(\mu\left(\bx_{\text{new}}\right),\sigma^{2}\left(\bx_{\text{new}}\right)\right)$, the GP posterior mean and variance at the data point $\bx_{\text{new}}$ can be computed as,
\begin{align*}
\mu\left(\bx_{\text{new}}\right)= & \mathbf{k}_{\text{new}} \left[ \mathbf{K} + \sigma^2 \idenmat \right]^{-1}\mathbf{y},\\
\sigma^{2}\left(\bx_{\text{new}}\right)= & k_{\text{new}}-\mathbf{k}_{\text{new}} \left[ \mathbf{K} + \sigma^2 \idenmat \right]^{-1} \mathbf{k}_{\text{new}}^{T}.
\end{align*}
As GPs give full uncertainty information with any prediction, they provide a flexible nonparametric prior for Bayesian optimization. We refer the interested readers to \cite{Rasmussen_2006gaussian} for further details on GPs.

\subsection{Experimental Setup}

\paragraph{Datasets} We evaluate our proposed framework using two real-world datasets from the UC Irvine (UCI) Data Repository\footnote{https://archive.ics.uci.edu}. The first dataset is the \textit{Adult} dataset \cite{Kohavi1996Adult}. Each record in this dataset describes the census data of a person (e.g., age, race, gender, relationship, workclass, education), and the goal is to predict whether the income of this person exceeds 50,000 USD per year. We choose the subgroup attributes to be age, race, gender and relationship. Note that for the age attribute, we group the values to be within 6 distinct values representing people with ages less than 20, from 20 to 30, ..., 50 to 60, and more than 60. The second dataset we use for evaluation is the \textit{Bike Sharing} dataset \cite{Fanaee2013BikeSharing}. This dataset contains the hourly count of rental bikes between years 2011 and 2012 in Capital bikeshare system with the corresponding weather and seasonal information. The goal is to predict the hourly count of rental bikes given a particular weather and seasonal information. For this dataset, we choose the subgroup attributes to be season, weather, hours, and working day. For the hour attribute, we group the values to be within 5 distinct values representing early morning, morning, afternoon, evening and midnight.

\paragraph{Machine Learning Systems} We split each dataset into two parts: training and testing. We have two choices of training size: 1000 and 2000 data points, the rest of the data is the testing part. We train different ML systems on the training parts and then use our proposed framework to find the worst-performing subgroup of these ML systems on the test part. In particular, for each subgroup $x$, we compute the metric value $M_x(C)$ by collecting the labelled data points in the dataset according to this subgroup. For classification problems (\textit{Adult} and \textit{Bank} datasets), we use the Logistic Regression and the Gradient Boosting as the ML systems. For the regression problem (\textit{Bike Sharing} dataset), we use the Linear Regression and Gradient Boosting as the ML systems. All the ML systems are implemented using sklearn \cite{scikit-learn2011} with default settings. For each ML system, we repeat the experiments 20 times with different split of training and testing parts.

\paragraph{Baselines} We compare our proposed framework with two baselines: Random Search and Exhaustive Search. With Random Search, we randomly select a number of subgroups, evaluate the ML system's performance associated with these subgroups, and then retrieve the worst-performing subgroup based on these performance values. With Exhaustive Search, we construct all the possible valid subgroups, and then sequentially evaluate the ML system's performance on each subgroup until the labelling budget is depleted. We then retrieve the worst-performing subgroup based on the evaluated performance values.
\end{document}